\documentclass[letterpaper, 10 pt, conference]{ieeeconf}
\usepackage{cite} 
\usepackage{multirow}
\usepackage[utf8]{inputenc}
\usepackage[misc]{ifsym}
\usepackage{graphicx}
\usepackage{tikz}
\usepackage{pifont}
\usepackage[misc]{ifsym}
\usepackage{tikz}
\usepackage{bbding}
\usepackage{pdfpages}
\usepackage{amsmath}
\usepackage[utf8]{inputenc}
\usetikzlibrary{arrows, shapes, positioning, shadows, trees, calc, decorations.markings}

\usepackage{amssymb}
\usepackage{hyperref}
\usepackage{booktabs}
\usepackage{algorithm}
\usepackage{algpseudocode}
\usepackage{balance}

\usepackage{color, colortbl}
\definecolor{LightCyan}{rgb}{0.88,1,1}

\hypersetup{
colorlinks=true,
linkcolor=black
}

\makeatletter
\newcommand{\printfnsymbol}[1]{%
  \textsuperscript{\@fnsymbol{#1}}%
}

\makeatother

\IEEEoverridecommandlockouts                

\title{\LARGE \bf
SurgVidLM: Towards Multi-grained Video Understanding with Large Language Model in Robot-assisted Surgery}

\author{Guankun Wang$^{1}$\printfnsymbol{1}\thanks{\printfnsymbol{1} Equal contribution}, Junyi Wang$^{1}$\printfnsymbol{1}, Wenjin Mo$^{2}$\printfnsymbol{1}, Long Bai$^{1}$\printfnsymbol{1}, Kun Yuan$^{3,4}$, Ming Hu$^{5}$, Jinlin Wu$^{6}$, \\
Junjun He$^{7}$, Yiming Huang$^{1}$, Nicolas Padoy$^{3}$, Zhen Lei$^{6}$, Hongbin Liu$^{6}$, Nassir Navab$^{4}$ and Hongliang Ren$^\dagger$$^{1}$
\thanks{$^{1}$ The Chinese University of Hong Kong, Hong Kong SAR, China.}
\thanks{$^{2}$ Sun Yat-sen University, Shenzhen, China.}
\thanks{$^{3}$ University of Strasbourg, Strasbourg, France.}
\thanks{$^{4}$ Technical University of Munich, Munich, Germany.}
\thanks{$^{5}$ Monash University, Melbourne, Australia.}
\thanks{$^{6}$ Centre for Artificial Intelligence and Robotics, HKISI-CAS, Hong Kong SAR, China.}
\thanks{$^{7}$ Shanghai AI Laboratory, Shanghai, China.}
}

\begin{document}
\maketitle
\thispagestyle{empty}
\pagestyle{empty}

\begin{abstract}
Surgical scene understanding is critical for surgical training and robotic decision-making in robot-assisted surgery. Recent advances in Multimodal Large Language Models (MLLMs) have demonstrated great potential for advancing scene perception in the medical domain, facilitating surgeons to understand surgical scenes and procedures. However, these methods are primarily oriented towards image-based analysis or global video understanding, overlooking the fine-grained video reasoning that is crucial for analyzing specific processes and capturing detailed task execution within a surgical procedure. To bridge this gap, we propose SurgVidLM, the first video language model designed to address both full and fine-grained surgical video comprehension. To train our SurgVidLM, we construct the SVU-31K that is a large-scale dataset with over 31K video-instruction pairs, enabling both holistic understanding and detailed analysis of surgical procedures. Building on this resource, SurgVidLM incorporates a two-stage StageFocus mechanism: the first stage extracts global procedural context, while the second stage performs high-frequency local analysis guided by temporal cues. We also develop the Multi-frequency Fusion Attention to effectively integrate low- and high-frequency visual tokens, ensuring the preservation of critical task-specific details. Experimental results demonstrate that SurgVidLM significantly outperforms state-of-the-art Vid-LLMs of comparable parameter scale in both full and fine-grained video understanding tasks, showcasing its superior capability in capturing the context of complex robot-assisted surgeries. Our code and dataset are publicly available at \href{https://github.com/gkw0010/SurgVidLM} {https://github.com/gkw0010/SurgVidLM}.

\end{abstract}

\section{Introduction}
\label{sec:1}
Surgical vision-language models have attracted considerable attention due to their advancements in surgical scene understanding, encompassing tasks such as anatomy recognition~\cite{zhou2023text,wang2024video}, instrument detection~\cite{bai2023surgical,bai2025surgical}, and workflow recognition~\cite{yuan2024advancing,seenivasan2023surgicalgpt}. These techniques demonstrate substantial potential in enhancing surgical training and robotic decision-making, contributing to the advancement of intelligent surgical robot systems. Meanwhile, the development of Multimodal Large Language Models (MLLMs) has led to various studies utilizing multimodal data from surgical procedures to perform image-level surgical scene understanding~\cite{wang2024surgical, jin2024surgical, li2024llava1, wang2023foundation,schmidgall2024gp, wang2025endochat}. Trained through medical datasets, MLLMs can query biomedical images and provide more flexible responses to medical inquiries based on their inherent reasoning ability. However, image-based surgical scene understanding methods primarily focus on static visual cues, limiting their ability to capture the complex temporal dependencies and procedural context inherent in surgical videos. To address this limitation, Video Large Language Models (Vid-LLMs) extend the reasoning and interaction capabilities of LLMs to video understanding by using parallel video-text datasets~\cite{yuan2024procedure}. By extracting temporal features, Vid-LLM can effectively perform joint modeling of temporal representations and text~\cite{anne2017localizing, gao2017tall}. For the training of Vid-LLMs, the acquisition of video-text datasets remains a significant challenge, particularly in the surgical domain, due to the requirement for costly expert annotations and specialized knowledge. Despite efforts to construct surgical video understanding data, existing datasets~\cite{li2024llava, jin2024surgical} are limited by accessibility or lack of reasoning annotations.

Recently, extensive progress has been made on video understanding in the computer vision field~\cite{chandrasegaran2025hourvideo,zhang2025videollama,lin2023video, li2023videochat, zhang2023video}. However, they face challenges when directly applied to the surgical scenario due to their reliance on general video datasets, which differ significantly from the complex, domain-specific context inherent in surgical procedures~\cite{chen2024vs}. In the surgical scenario, there are some efforts on the construction of large-scale video-instruction datasets and the integration of image and video visual representations for improved video understanding~\cite{li2024llava, jin2024surgical}. However, these approaches remain constrained by limited data accessibility or the reliance on image-based annotations for fine-tuning. Additionally, surgical procedures are inherently multi-stage and dynamic, involving a sequence of complex, interdependent steps~\cite{bai2024ossar}. General video understanding methods that focus on full video understanding have difficulty understanding multi-step interactions and reasoning over visual and textual knowledge. To address this, datasets with multi-grained video-text annotations are essential for enabling Vid-LLMs to perform precise surgical video reasoning. 
Due to computational resource constraints, Vid-LLMs extract visual information by low-frequency sampling of video frames, which inevitably results in the loss of critical local information essential for fine-grained video understanding~\cite{nie2024slowfocus}. If Vid-LLMs can locate the time of video clips related to questions and sample frames at a higher frequency, specifically, the visual information will be better extracted and preserved, thereby facilitating more precise fine-grained analysis.

To address these challenges in surgical video understanding, we first construct SVU-31K, a large-scale multi-grained surgical video dataset containing 31K video-text pairs. Based on SVU-31K, we propose the SurgVidLM, a novel video-language model that progressively refines video understanding from global to local contexts. SurgVidLM effectively integrates multi-frequency visual features, enabling the preservation of essential procedural cues critical for surgical video analysis. Specifically, our contributions are as follows:
\begin{itemize}
\item[--] We propose \textbf{SurgVidLM}, the first multi-grained surgical video language model tailored for robot-assisted surgery, which supports both holistic full video comprehension and fine-grained visual reasoning. 
\item[--] We develop \textbf{SVU-31K}, a large-scale surgical video understanding dataset annotated through a novel knowledge augmentation pipeline. SVU-31K provides both full video and fine-grained video understanding task annotations, which allow the model to achieve structured and context-aware surgical video comprehension.
\item[--] SurgVidLM incorporates the \textbf{StageFocus mechanism}, which progressively refines global-to-local video understanding. Additionally, \textbf{Multi-frequency Fusion Attention} is integrated to facilitate the interaction of low- and high-frequency visual tokens to preserve both contextual and detailed information. 
\item[--] Extensive experiments and ablation studies on SVU-31K demonstrate that SurgVidLM outperforms state-of-the-art Vid-LLMs of comparable parameter scale in multi-grained surgical video comprehension. These results highlight its potential for surgical scene understanding in robot-assisted surgery.
\end{itemize}

\section{Related Work}
\subsection{Video Large Language Models in the Computer Vision}
The scope of large language models (LLMs) has expanded beyond static images to encompass dynamic visual streams, giving rise to Video Large Language Models (Vid-LLMs)~\cite{chandrasegaran2025hourvideo,zhang2025videollama}. These models aim to combine the reasoning capability of LLMs with the temporal and multimodal complexity of videos. 
Early attempts such as VideoChat~\cite{li2023videochat} focus on enhancing interactive understanding by fine-tuning LLMs with video-centric instruction data, while Video-LLaMA~\cite{zhang2023video} trains visual-language and audio-language branches, enabling a deeper understanding of both visual and auditory content. Video-LLaVA~\cite{lin2023video} aligns projections to jointly train on images and videos, learning unified visual representations. More recently, VideoLLaMA~\cite{cheng2024videollama} enhances the perception of temporal dynamics by incorporating a spatial-temporal convolutional connector. VTimeLLM~\cite{huang2024vtimellm} introduces prefixing frame information to inject temporal awareness. TimeChat~\cite{ren2024timechat} combines time-stamped instructions with the query before attention computation, enabling finer alignment between temporal context and language understanding. Although existing Vid-LLMs have demonstrated remarkable video understanding capabilities, they still struggle in the surgical domain. Their training mainly relies on open-domain video corpora, lacking visual details like subtle tool–tissue interactions, anatomical structures with high visual similarity, and procedural workflows that unfold under strict temporal constraints.

\subsection{Surgical Multimodal Large Language Models}
The integration of MLLMs into the medical community has gained significant momentum. Notably,  GP-VLS~\cite{schmidgall2024gp} develops a general-purpose vision-language model for surgery that integrates medical and surgical knowledge with visual understanding. S\textsuperscript{2}Can~\cite{hou2024memory} proposes a memory-augmented multimodal LLM that uses self-contained inquiry to generate direct and indirect memory cues, achieving excellent performance on surgical VQA tasks. In addition, Endochat~\cite{wang2026endochat} presents a vision-language foundation model trained on a large-scale multimodal surgical database. Building on these advances, researchers have further explored the application of MLLMs in surgical video understanding. LLaVA-Surg~\cite{li2024llava} proposes the Surg-QA dataset, comprising 102K video-instruction pairs. However, Surg-QA is not publicly available. Although Surgical-LLaVA~\cite{jin2024surgical} achieves the integration of image and video visual representations for video understanding, it is fine-tuned using image-based annotations. In this paper, we introduce a novel, large-scale surgical video understanding dataset and a multi-grained video large language model to mitigate above problems.

\label{sec:2}
\begin{figure*}[]
    \centering
    \includegraphics[width=0.9\linewidth]{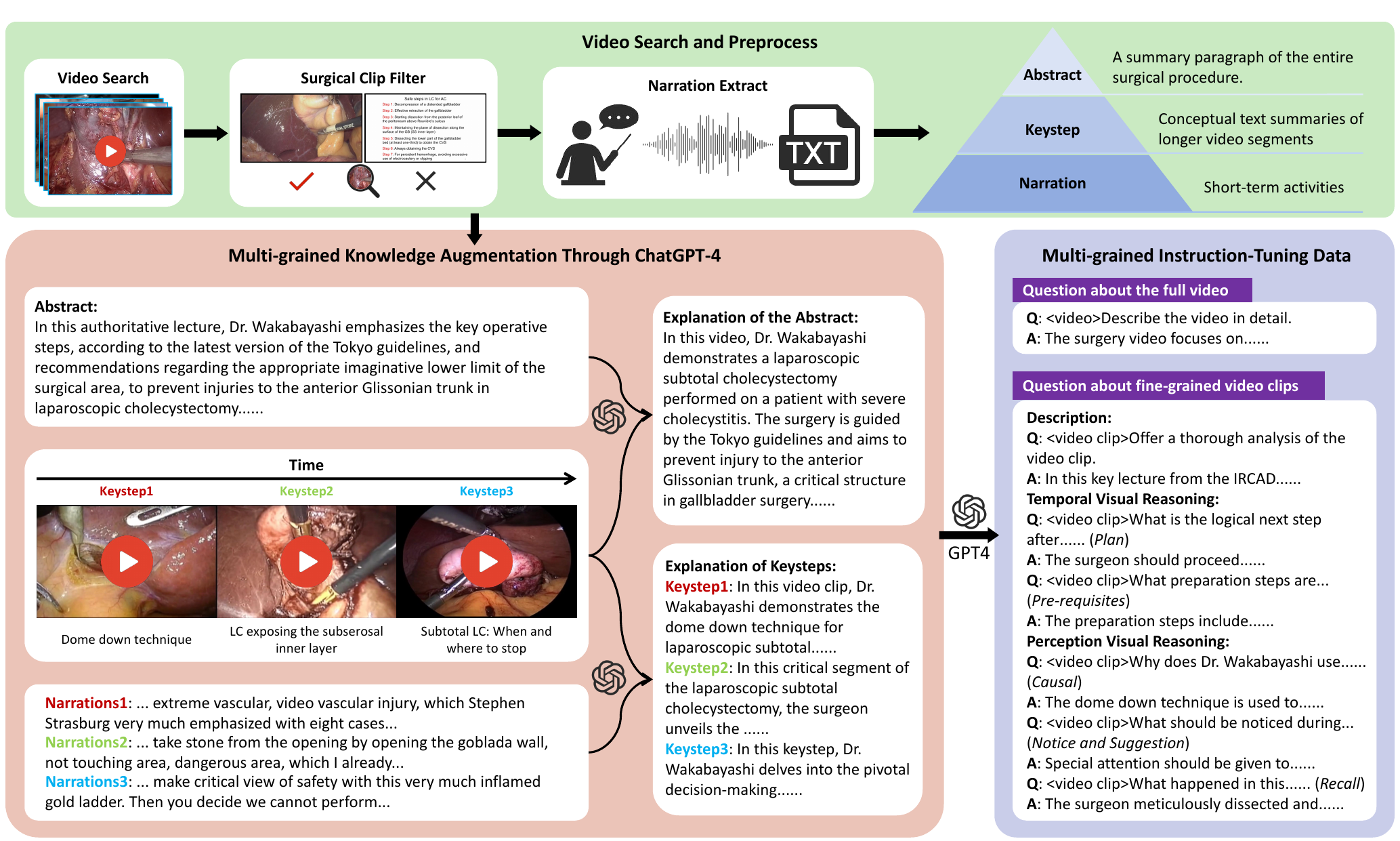}
    \caption{Data collection and construction pipeline for SVU-31K.}
    \label{fig:dataset}
\end{figure*}

\section{Methodology}
\subsection{SVU-31K Dataset Construction}
\label{sec:2.1}
Beyond the scarcity of benchmarks for surgical video understanding, existing studies~\cite{li2024llava, twinanda2016endonet} do not investigate fine-grained video understanding. To bridge this gap, we present SVU-31K, a high-quality instruction-tuning dataset tailored to endow Vid-LLMs to understand both full and fine-grained surgical videos. The construction pipeline comprises the following three steps, which are shown in Fig.~\ref{fig:dataset}.


\subsubsection{Video Search and Preprocess}
We collect 1882 robot-assisted surgical videos from YouTube and WebSurg, encompassing a range of surgical specialties: Pancreatic Surgery, Digestive Surgery, Hernia Surgery, and Colorectal Surgery, etc. Each video is accompanied by its abstract text, as well as keystep annotations with corresponding timecodes. To structure the data, we segment each video into clips based on the annotated key steps. A pre-trained ResNet-50 model is then employed to filter out non-surgical segments, ensuring that only clips containing surgical content are retained. Specifically, the ResNet-50 model is trained with images extracted from 200 videos and achieves 94\% classification accuracy. During the segments classification, we sample at 1 fps and classify the segment as surgical if at least 80\% of the frames are predicted as surgical—this yields 100\% accuracy for the segment filter. The audio from the remaining clips is then transcribed into narrations using OpenAI Whisper~\cite{yuan2023learning}, followed by GPT-4 to correct spelling and semantic errors. As a result, we extract a three-level hierarchical knowledge representation comprising Abstract, Keystep, and Narration. These three levels of knowledge—Abstract providing a global overview, Keystep detailing key phases of the surgery, and Narration capturing short-term surgical activities—offer a comprehensive representation of the surgical process. With them, we can achieve a nuanced understanding of surgical tasks at varying levels of granularity.

\subsubsection{Multi-grained Knowledge Augmentation}
While texts at varying levels provide diverse insights into surgical videos, there are still some limitations. Specifically, Abstracts often contain redundant content. Keysteps, while concise and structured, are limited in linguistic diversity and lack detailed contextual information. Meanwhile, narratives, extracted from audio transcriptions, tend to be fragmented and incoherent. These incomplete, noisy, or overly concise textual inputs can lead to suboptimal question generation, ultimately limiting the model’s ability to reason about surgical procedures effectively. To address this, we introduce multi-grained knowledge augmentation to enhance textual quality across all levels. First, we employ GPT-4 to reorganize and supplement the abstract texts based on keysteps, removing extraneous details. Then, keystep texts are enriched with narrative details, adding procedural, anatomical, and instrumental context. We integrate narration texts into the keystep texts rather than expanding on them, as such clips are too brief to be used directly. This multi-grained augmentation yields a contextually rich textual representation, which serves as a structured foundation for the following instruction-tuning.

\subsubsection{Multi-grained Instruction-Tuning Data}
Using the augmented texts as context, we construct a multi-grained instruction-tuning dataset with GPT-4, introducing granularity-specific visual question-answer pairs to enhance the model’s understanding of surgical videos. At the full-video level, the model is trained on summarization and perception tasks. For keystep-based video clips, we design three types of video understanding tasks, as illustrated in the purple part of Fig.~\ref{fig:dataset}: (i) \textit{Summary description task} requires the model to generate a concise summary of surgical keysteps. (ii) \textit{Temporal visual reasoning task (Plan, Pre-requisites)} intends to understand the temporal order of surgical operations and infer the prerequisites or planning steps for specific actions. For example, if a clip shows the suturing process, the task might require the model to deduce that the prerequisite steps of tissue alignment and needle preparation are needed prior to suturing. (iii) \textit{Perceptual visual reasoning task (Causal, Notice and Suggestion, Recall)} infers about causal relationships, suggestions, and recalls key information during surgical procedures. When observing a surgeon encountering difficulty with tissue manipulation, the model should analyze the causal relationship between improper force application and tissue damage, suggest a correction, and recall prior successful approaches in similar scenarios. Each keystep-based QA pair is associated with a timecode, enabling the model to align the QA pair with a specific video clip when extracting visual information.
Finally, the SVU-31K dataset includes 1.8K QA pairs for the full video and 29.4K QA pairs for fine-grained video understanding tasks. Detailed information regarding the text formats, prompt designs and the number of each QA task is provided in the supplement. 

\begin{figure*}[]
    \centering
    \includegraphics[width=0.84\linewidth]{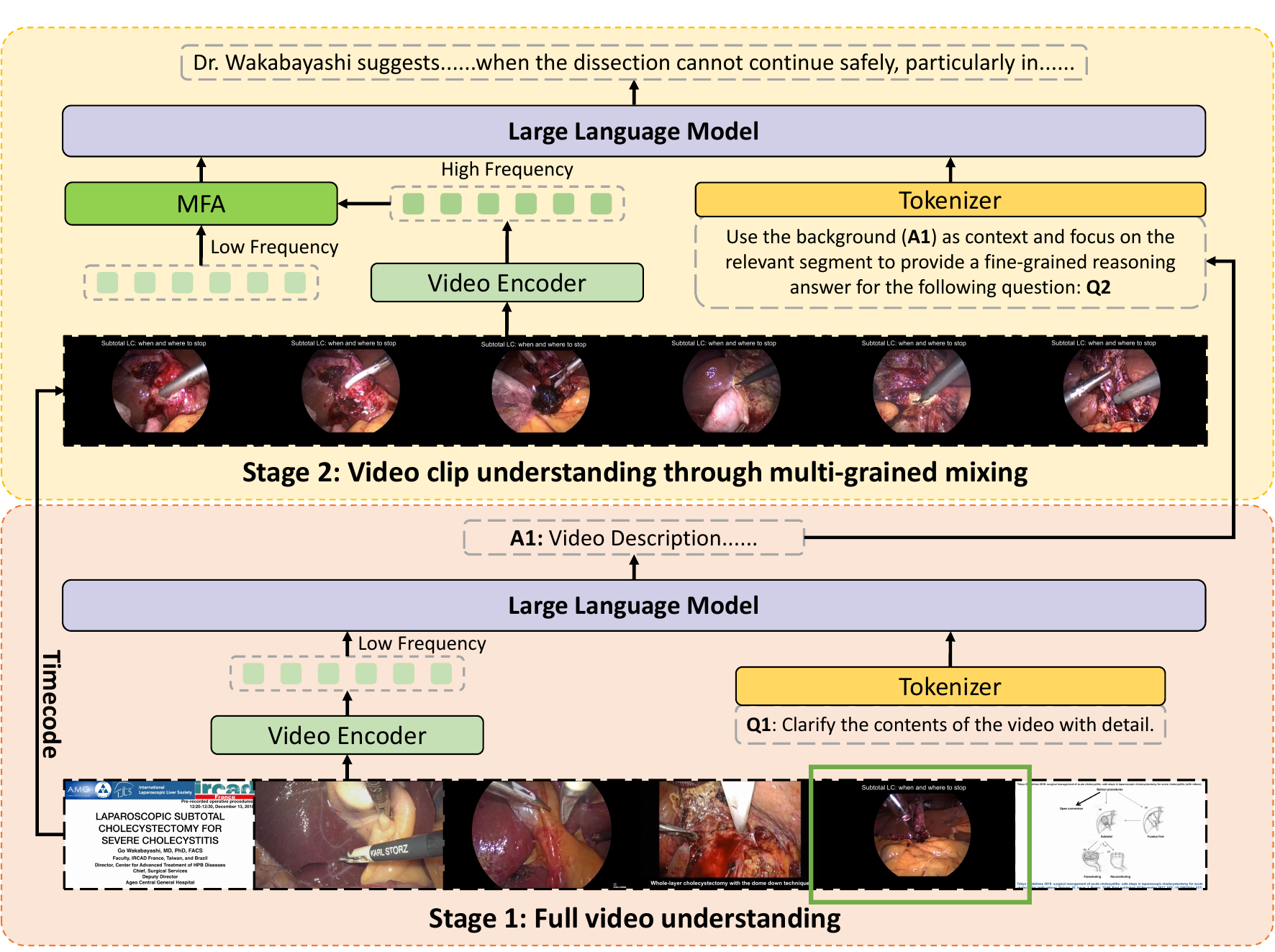}
    \caption{The architecture of SurgVidLM. Stage 1 focuses on a holistic understanding of the full video, while Stage 2 fuses the information from both the current and previous stages to achieve accurate, fine-grained video understanding.}
    \label{fig:main_figure}
\end{figure*}

\subsection{SurgVidLM}
\label{sec:2.2}
\subsubsection{Preliminaries}
We develop the SurgVidLM upon the Qwen2-VL framework, which integrates vision encoders and language model Qwen2\cite{wang2024qwen2}. Let \( V = \{V(t) \in \mathbb{R}^{H \times W \times 3} \mid t = 0, \dots, T\} \) represents a video sequence comprising \( T \) frames, along with an associated query \( q \). The visual encoder extracts frame-level features from downsampled frames \( V_F \), which are sparsely sampled at fixed intervals \( I_F \), yielding low-frequency visual tokens \( X_F \). Subsequently, \( X_F \) is transformed into the language embedding space via the visual adapter and concatenated with language tokens for LLM response generation.

\subsubsection{StageFocus for Multi-grained Video Understanding}
While low-frequency sampling has proven effective for general video understanding tasks and can maintain computational efficiency, it faces limitations in handling surgical video understanding, where fine-grained procedural cues often occur within short and subtle intervals. To address these challenges, we introduce the StageFocus mechanism that adopts two-stage reasoning to perform the multi-grained, progressive understanding of surgical videos. Specifically, in Stage 1, SurgVidLM follows the original Qwen2-VL architecture to conduct a holistic analysis of the entire video, exemplified by prompts like “\textit{\texttt{<video>}\textbackslash nClarify the contents
of the video with detail.}” This broad understanding serves as the foundation for deeper, more specific interpretations in the following stage. In Stage 2, SurgVidLM is provided with question-related video clips' timecode and shifts its focus to fine-grained video analysis. High-frequency sampling is performed on the fine-grained video clips with the interval $I_C$. The sampled frames are subsequently processed by the visual encoder, producing the high-frequency visual token  \(X_C\), which is combined with \( X_F \) for fine-grained video understanding and reasoning. For the input text, we use the answer from the first stage as the background knowledge, such as “\textit{\texttt{<video>}Use the background (\texttt{<Answer1>}) as context and focus on the relevant segment to provide a fine-grained reasoning answer for the following question:\textbackslash n \texttt{<Question2>}}”. By integrating the mixed-frequency visual tokens with the stage-one textual context, SurgVidLM derives the final answer, which balances detailed temporal observations with comprehensive video-level semantics, thereby mitigating the information inaccuracy and hallucinations caused by information loss.

\subsubsection{Multi-frequency Fusion Attention}
Surgical videos exhibit strong micro-level detail within each surgical step and macro-level temporal dependencies across surgical steps. Therefore, effective fine-grained surgical video understanding requires capturing the interactions among multiple surgical steps, necessitating the integration of global context with local features. Due to the inherent coarseness of the low-frequency visual tokens and the detailed specificity of high-frequency visual tokens, directly concatenating them for input into the language model may result in suboptimal performance~\cite{nie2024slowfocus}.  Consequently, we introduce the Multi-frequency Fusion Attention (MFA) in Stage 2: 
\begin{equation}
    E\left(X_C, X_F\right)=softmax\left(\frac{X_CX_F^{T}}{\sqrt{d}}\right) X_F
\end{equation}
\begin{equation}
    X_{fuse} = X_C + E\left(X_C, X_F\right)
\end{equation}

As shown in Eq. (1), high-frequency visual token \(X_C\) serves as queries, while low-frequency visual token \(X_F\) serves as keys and values in a cross-attention mechanism to obtain aggregated global context conditioned on \(X_C\). Then the residual connection is applied to add the aggregated global context back to the high-frequency tokens to obtain context-enhanced representation \(X_{fuse}\), as shown in Eq. (2).  Finally, \(X_{fuse}\) is fed into the LLM as visual input. MFA enriches high-frequency visual tokens with global context aggregated from low-frequency visual tokens, aligning transient surgical cues with the broader procedural state and thereby enhancing temporal coherence. In this way, MFA achieves a balanced fusion that preserves fine-grained surgical details while grounding them within the overall procedural context. Therefore, SurgVidLM can capture the nuanced interactions within surgical scenes, generating precise and contextually relevant insights for fine-grained video understanding.



\subsubsection{Training Strategy}
Following the StageFocus mechanism, we design a two-phase training strategy that endows SurgVidLM with the capacity for multi-grained surgical video understanding. The first phase is dedicated to knowledge augmentation, where the model is trained for a holistic understanding of long surgical video sequences. In this phase, we utilize QA pairs corresponding to full videos from the SVU-31K dataset. The language model (LLM) is fine-tuned using LoRA~\cite{hu2021lora}. The second phase focuses on enhancing fine-grained surgical video understanding and MFA integration. Under the StageFocus mechanism, we train the LLM and MFA modules with the fine-grained part of the SVU-31K dataset, facilitating cross-stage information fusion. The visual encoder's parameters are frozen in both stages.

\begin{table*}[t]
\vspace{0.2cm}
\caption{Comparison of SurgVidLM with zero-shot Vid-LLMs across multi-grained video understanding tasks on the SVU-31K dataset. CI, DO, CU and TU means Correctness of Information, Detail Orientation, Contextual Understanding and Temporal Understanding, respectively.}
\label{main_table}
\resizebox{\textwidth}{!}{
\begin{tabular}{c|c|cccc|cccc}
\toprule[1pt]
\multirow{2}{*}{Model} & \multirow{2}{*}{Finetuning} & \multicolumn{4}{c|}{Full Video Description}                                                & \multicolumn{4}{c}{Fine-grained Video Description}                                        \\ \cline{3-10} 
                       &                             & CI                   & DO                   & CU                   & TU                    & CI                   & DO                   & CU                   & TU                   \\ \hline
Video-LLaVA-7B~\cite{lin2023video}            & \scriptsize{\XSolidBrush}       &                   1.03	&1.01	&1.01	&1.03	&1.15	&1.09	&1.24	&1.21                      \\
GroundingGPT-7B~\cite{li2024groundinggpt}           & \scriptsize{\XSolidBrush}                           &                      1.05	&1.06	&1.15	&1.08	&1.05&	1.16	&1.26	&1.16 \\
VideoLLaMA3-7B~\cite{zhang2025videollama}            & \scriptsize{\XSolidBrush}                           &                      1.66	&1.71&	2.21	&1.86	&1.58&	1.87&	2.29	&1.95                      \\
Qwen2-VL-7B~\cite{wang2024qwen2}             & \scriptsize{\XSolidBrush}                           &                      1.59	&1.27	&1.36	&1.85	&1.69	&1.54	&1.42	&1.98                      \\
Qwen2.5-VL-7B~\cite{bai2025qwen2}               & \scriptsize{\XSolidBrush}                           &                      1.22	&1.09&	1.64	&1.53&	1.82&	1.77&	2.49	&1.95                      \\
 \hline
VideoLLaMA3-7B~\cite{zhang2025videollama}            & \textbf{\scriptsize{\CheckmarkBold}}                  &                      1.91	&2.05	&\textbf{2.79}	&2.21	&1.66	&1.98	&2.58	&2.21                      \\
Qwen2-VL-7B~\cite{wang2024qwen2}               & \textbf{\scriptsize{\CheckmarkBold}}                  &                      2.40	&2.11	&2.69	&2.22&	1.78	&1.75&	1.98	&2.23                     \\
Qwen2.5-VL-7B~\cite{bai2025qwen2}              & \textbf{\scriptsize{\CheckmarkBold}}                  & 2.28	&\textbf{2.32}	&2.76&	2.14&	1.94&	2.15&	2.66	&2.17 \\
SurgVidLM                   & \textbf{\scriptsize{\CheckmarkBold}}                  &  \textbf{2.40}&	2.11	&2.69&	\textbf{2.22}	&\textbf{2.27}	&\textbf{2.62}	&\textbf{3.06}	&\textbf{2.44} \\ \hline
\multirow{2}{*}{Model} & \multirow{2}{*}{Finetuning} & \multicolumn{4}{c|}{Fine-grained Temporal Visual Reasoning}                                & \multicolumn{4}{c}{Fine-grained Perception Visual Reasoning}                              \\ \cline{3-10} 
                       &                             & BLEU-4               & CIDEr                & METEOR               & ROUGE-L               & BLEU-4               & CIDEr                & METEOR               & ROUGE-L              \\ \hline
Video-LLaVA-7B~\cite{lin2023video}            & \scriptsize{\XSolidBrush}       & 5.65&	2.17	&11.59	&19.73&	10.76	&5.48&	13.52	&25.93\\
GroundingGPT-7B~\cite{li2024groundinggpt}          & \scriptsize{\XSolidBrush}                           &                      2.81	&1.44	&10.52	&17.86&	4.47	&1.51	&11.96	&21.34                    \\
VideoLLaMA3-7B~\cite{zhang2025videollama}            & \scriptsize{\XSolidBrush}                           &                      7.68&	1.92	&11.53&	23.19&	9.57&	3.97&	12.63	&24.21    \\
Qwen2-VL-7B\cite{wang2024qwen2}               & \scriptsize{\XSolidBrush}                           &                      5.18	&1.66	&13.31&	22.9	&8.96	&2.47	&16.22	&26.51      \\
Qwen2.5-VL-7B~\cite{bai2025qwen2}             & \scriptsize{\XSolidBrush}                           & 4.41	&1.55&	13.09&	16.27	&3.19	&2.18	&15.04	&25.55            \\ \hline
VideoLLaMA3-7B~\cite{zhang2025videollama}            & \textbf{\scriptsize{\CheckmarkBold}}                  & 8.44&	2.88&	12.59&	28.7&	16.61	&7.94&	13.59	&32.18            \\
Qwen2-VL-7B~\cite{wang2024qwen2}               & \textbf{\scriptsize{\CheckmarkBold}}                  & 9.74	&2.97	&13.40	&29.76	&15.48	&8.83	&16.87&	34.48            \\
Qwen2.5-VL-7B~\cite{bai2025qwen2}             & \textbf{\scriptsize{\CheckmarkBold}}                  & 7.45&	2.96	&13.49&	20.63	&6.18	&6.78&	16.74	&33.34            \\
SurgVidLM                   & \textbf{\scriptsize{\CheckmarkBold}}                  & \textbf{10.10}&	\textbf{3.83}&	\textbf{14.13}	&\textbf{31.58}&	\textbf{16.71}&	\textbf{9.76}&	\textbf{18.27}	&\textbf{37.51} \\ \bottomrule[1pt]
\end{tabular}}
\end{table*}

\section{Experiments}
\label{sec:3}

\subsection{Implementation Details}
We adopt the pretrained Qwen2-VL-7B-Instruct~\cite{wang2024qwen2} as the initial weights of our SurgVidLM. In the first training phase, we use a batch size of 4, while in the second phase, the batch size is increased to 16. Optimization is performed using AdamW~\cite{loshchilov2017decoupled} with a learning rate of $1\times10^{-4}$, and the model is trained for 1 epoch in total. The low-frequency sampling interval $I_F$ is set to one frame per second, whereas the high-frequency sampling interval $I_C$ is fixed at two frames per second. The resolution of videos is 224 $\times$ 224. The training process is conducted on 6 NVIDIA H100 GPUs, with the first phase taking approximately 3 hours and the second phase requiring around 15 hours. For comparison, we evaluate SurgVidLM against Video-LLaVA~\cite{lin2023video}, GroundingGPT~\cite{li2024groundinggpt}, VideoLLaMA 3~\cite{zhang2025videollama}, Qwen2-VL~\cite{wang2024qwen2} and Qwen2.5-VL~\cite{bai2025qwen2}. To provide a thorough evaluation of video description, we follow the evaluation pipeline in ~\cite{Maaz2023VideoChatGPT}, employing the GPT-4 model to assess the responses on a scale from 1 to 5 across four key aspects: correctness of information, detail orientation, contextual understanding, and Temporal Understanding. For the reasoning tasks, we utilize standard quantitative metrics, including BLEU-4, METEOR, ROUGE-L (scaled from 0 to 1 and presented as percentages in tables), and CIDEr (scaled from 1 to 10).


\subsection{Results}
Table~\ref{main_table} presents the comparison of our SurgVidLM against several state-of-the-art Vid-LLMs of comparable parameter scale across full-video description and fine-grained video understanding tasks. For full-video description, SurgVidLM achieves competitive performance and ties with Qwen2-VL, as our framework is specifically optimized for fine-grained analysis. All models exhibit substantial improvements after fine-tuning, demonstrating the effectiveness of the instruction-tuning process on the SVU-31K dataset. By contrast, for fine-grained video description, SurgVidLM demonstrates clear advantages. Compared with Qwen2-VL, it improves by +0.35 in CI, +0.74 in DO, +1.08 in CU, and +0.08 in TU, indicating that the proposed StageFocus mechanism combined with MFA enables more precise complex surgical scene narratives. Similar trends are observed in reasoning tasks. For temporal reasoning, SurgVidLM outperforms all baselines, achieving gains of +0.36 BLEU-4, +0.86 CIDEr, +0.73 METEOR, and +1.82 ROUGE-L, validating the benefit of MFA in aligning short-term cues with long-range context. In perceptual reasoning, the margin is even more pronounced: +1.23 BLEU-4, +0.93 CIDEr, +1.40 METEOR, and +4.03 ROUGE-L, highlighting the role of fine-grained instruction-tuning (e.g., Causal, Recall) in promoting procedure-specific reasoning rather than generic observations. It is noteworthy that, while Qwen2.5-VL demonstrates marginal advantages in full-video description, Qwen2-VL achieves a more balanced performance across both full-video and fine-grained video understanding tasks. Consequently, SurgVidLM is built upon the Qwen2-VL model.

\begin{figure*}[t]
    \centering
    \includegraphics[width=\linewidth, trim=0 25 0 0]{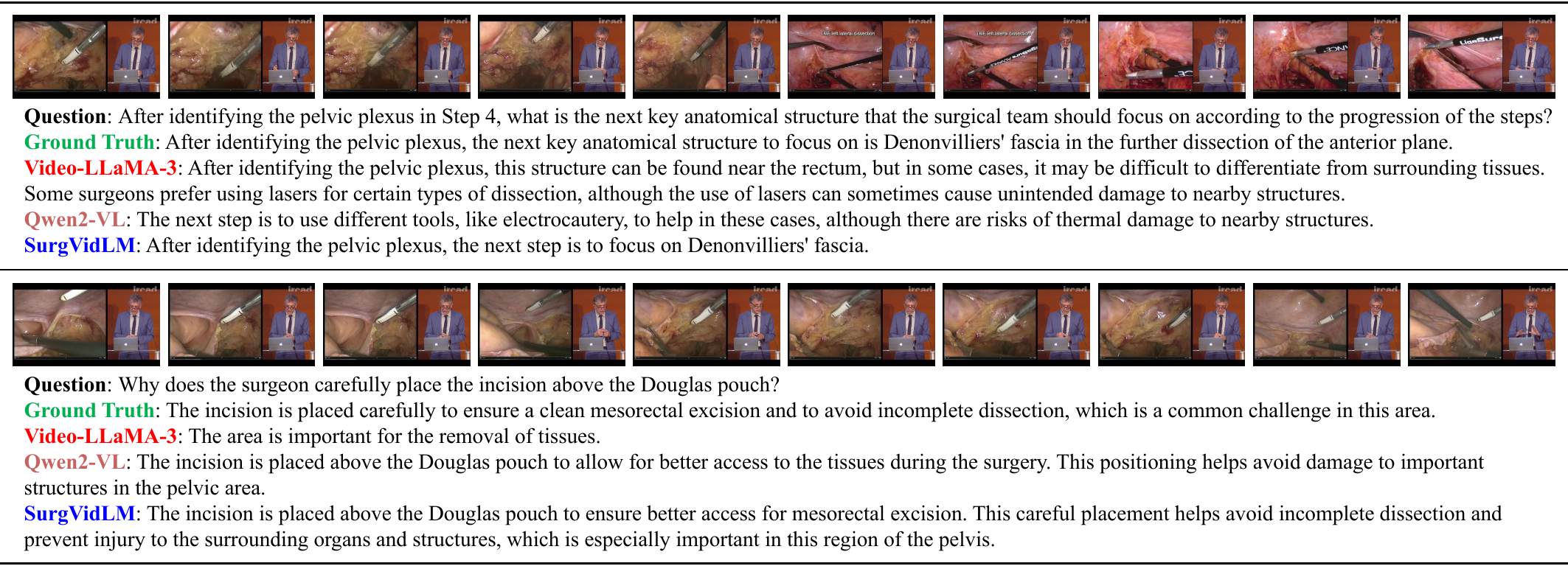}
    \caption{Qualitative examples of fine-grained video reasoning tasks.}
    \label{fig:demo}
\end{figure*}

Fig.~\ref{fig:demo} highlights SurgVidLM’s superior performance in procedural reasoning and anatomical understanding. When identifying the pelvic plexus, SurgVidLM correctly infers that the subsequent key anatomical landmark is Denonvilliers’ fascia, closely aligning with the surgical ground truth. In contrast, VideoLLaMA-3 and Qwen2-VL tend to produce vague or tool-oriented responses, failing to capture the precise procedural progression. Similarly, when asked about the rationale for placing the incision above the Douglas pouch, SurgVidLM not only provides the correct anatomical reasoning—ensuring complete mesorectal excision and avoiding incomplete dissection—but also articulates the clinical motivation behind this choice. These qualitative results demonstrate that, beyond surface-level descriptions, SurgVidLM is capable of producing step-wise surgical objectives (mesorectal excision) and risks (avoiding tissue injury), demonstrating the effectiveness of instruction-tuning tasks like Plan, Pre-requisites, Causal, Notice, and Suggestion.

Our ablation study confirms the complementary roles of StageFocus and Multi-frequency Fusion Attention, with their combination achieving the best performance across video description and visual reasoning tasks (Table~\ref{ablation}). When only StageFocus is applied, the model already shows notable gains in correctness and temporal coherence by progressively refining global-to-local reasoning. However, this configuration still suffers from limited integration of global context into fine-grained observations, leading to incomplete detail preservation. The joint application of StageFocus and MFA achieves the best performance across all metrics: StageFocus ensures progressive refinement of surgical procedure comprehension, while MFA effectively bridges global and local representations, together enabling SurgVidLM to deliver an accurate, detail-oriented, and contextually coherent understanding of surgical videos.

\begin{table*}[t]
\vspace{0.2cm}
\caption{Ablation study on StageFocus and Multi-grained
Fusion Attention.}
\resizebox{\textwidth}{!}{
\begin{tabular}{cc|cccc|cccc}
\toprule[1pt]
\multirow{2}{*}{StageFocus} & \multirow{2}{*}{\begin{tabular}[c]{@{}c@{}}Multi-grained\\  Fusion Attention\end{tabular}} & \multicolumn{4}{c|}{Fine-grained Video Description}           & \multicolumn{4}{c}{Fine-grained Visual Reasoning} \\ \cline{3-10} 
                            &                                                                                            & CI           & DO            & CU      & TU       & BLEU-4     & CIDEr     & METEOR     & ROUGE-L     \\ \hline
\scriptsize{\XSolidBrush}                    & \scriptsize{\XSolidBrush}                                                                                   & 1.78	&1.75&	1.98	&1.63&	15.01&	8.47&	16.57&	33.06       \\
\scriptsize{\CheckmarkBold}   & \scriptsize{\XSolidBrush}                                                                                   &2.11	&2.37	&3.04&	2.23	&16.08	&9.17&	17.78	&36.95       \\ 
\scriptsize{\CheckmarkBold}   & \scriptsize{\CheckmarkBold}                                                                  & \textbf{2.27}	&\textbf{2.62}	&\textbf{3.06}	&\textbf{2.44} & \textbf{16.16}      & \textbf{9.30}    & \textbf{17.91}      & \textbf{36.97}       \\ \bottomrule[1pt]
\end{tabular}}
\label{ablation}
\end{table*}

\begin{table*}[t]
\centering
\caption{Ablation study on sampling frequency (fps).}
\resizebox{0.9\textwidth}{!}{
\begin{tabular}{c|cccc|cccc}
\toprule[1pt]
\multirow{2}{*}{\begin{tabular}[c]{@{}c@{}}Sampling\\ Frequency\end{tabular}} & \multicolumn{4}{c|}{Fine-grained Video Description}                                                & \multicolumn{4}{c}{Fine-grained Visual Reasoning}                                                                 \\ \cline{2-9} 
                                                                              & \multicolumn{1}{c}{CI} & \multicolumn{1}{c}{DO} & \multicolumn{1}{c}{CU} & \multicolumn{1}{c|}{TU} & \multicolumn{1}{c}{BLEU-4} & \multicolumn{1}{c}{CIDEr} & \multicolumn{1}{c}{METEOR} & \multicolumn{1}{c}{ROUGE-L} \\ \hline
0.5                                                                             &                        2.14&	2.06	&2.52&	2.02	&14.52	&8.52	&17.79	&35.71        \\
1                                                                             &                        2.23	&2.48	&2.84&	2.29	&16.1	&9.21	&17.88	&35.98                            \\
2                                                                             &                        2.27 & \textbf{2.62} & \textbf{3.06} & 2.44 & 16.16      & \textbf{9.30}    & \textbf{17.91}      & 36.97                               \\
4                                                                             &                        \textbf{2.33}	&2.59	&2.96	&\textbf{2.62}	&\textbf{16.19}	&9.29	&17.90	&\textbf{36.99}                           \\ \bottomrule[1pt]
\end{tabular}}
\label{ablation_freq}
\end{table*}
We also investigate the influence of different sampling frequencies on model performance, which is illustrated in Table~\ref{ablation_freq}. Notably, the results exhibit substantial improvements when increasing the sampling frequency from 0.5 fps to 1 fps, with further gains from 1 fps to 2 fps, but no obvious improvements from 2 fps to 4 fps. We attribute this to two primary factors. Firstly, when the sampling rate increases from a lower value, the newly sampled frames are temporally farther from their neighbors, and therefore can introduce considerable incremental information. As the sampling frequency continues to increase from a higher value, the newly sampled frames become increasingly similar to the adjacent frames, yielding limited additional information while increasing redundancy. Second, surgical videos contain transient noises, such as camera shake and motion artifacts from instruments, and the higher sampling rates increase the probability of capturing these noises, diminishing the benefit of denser sampling. Therefore, continuously increasing sampling frequency does not produce linear improvements while increasing the computational costs. To strike a balance between capturing detailed information and computational efficiency, we use 2 fps as the sampling frequency for fine-grained videos as an optimal trade-off.

\section{Conclusions}
\label{sec:4}
In this work, we introduce SVU-31K, a large-scale multi-grained surgical video understanding dataset that comprises over 31K video-instruction pairs. Based on SVU-31K, we propose SurgVidLM, the first surgical video language model specifically designed for multi-grained surgical video comprehension. Our proposed StageFocus mechanism utilizes a two-stage reasoning approach to guide the model to analyze surgical procedures from holistic to detail progressively. The integration of Multi-frequency Fusion Attention further enhances the model's performance by effectively combining low- and high-frequency visual tokens, thus preserving critical information essential for fine-grained surgical scene understanding. Experimental results demonstrate that SurgVidLM significantly outperforms SOTA Vid-LLMs across various multi-grained surgical video understanding tasks. These findings underscore the potential of SurgVidLM to advance the field of surgical video understanding. Future work will focus on expanding the dataset and exploring the application of SurgVidLM in real-time surgical environments, thereby contributing to improved clinical outcomes through enhanced video analysis capabilities.
\balance
\bibliographystyle{IEEEtran}
\bibliography{mybib}

\end{document}